\def\cvprPaperID{11508}
\def\confName{ICCV}
\def\confYear{2023}

\def\paperTitle{Real Face Foundation Representation Learning \\ for Generalized Deepfake Detection}

\def\authorBlock{
    Liang Shi$^{1,2}$ \qquad
    Jie Zhang$^{1,2}$ \qquad
    Shiguang Shan$^{1,2}$ \\
    $^1$ Institute of Computing Technology, Chinese Academy of Sciences \\
    $^2$ University of Chinese Academy of Sciences \\
    {\tt\small liang.shi@vipl.ict.ac.cn;  \{zhangjie, sgshan\}@ict.ac.cn}
}

\newif\ifreview 
\newif\ifarxiv \newcommand{\arxiv}{\arxivtrue}
\newif\ifcamera 
\newif\ifrebuttal 

\arxiv % \review OR \arxiv OR \cameraready

\pdfoutput=1
\documentclass[10pt,twocolumn,letterpaper]{article}
\ifreview \usepackage[review]{cvpr} \fi
\ifarxiv \usepackage[pagenumbers]{cvpr} \fi
\ifrebuttal \usepackage[rebuttal]{cvpr} \fi
\ifcamera \usepackage{cvpr} \fi

\usepackage{graphicx}
\usepackage{amsmath}
\usepackage{amssymb}
\usepackage{booktabs}

\usepackage{times}
\usepackage{microtype}
\usepackage{epsfig}
\usepackage[table,xcdraw]{xcolor}
\usepackage{caption}
\usepackage{float}
\usepackage{placeins}
\usepackage{color, colortbl}
\usepackage{stfloats}
\usepackage{enumitem}
\usepackage{tabularx}
\usepackage{xstring}
\usepackage{multirow}
\usepackage{xspace}
\usepackage{url}
\usepackage{subcaption}
\usepackage{xcolor}
\usepackage[hang,flushmargin]{footmisc}
\usepackage{xcolor,colortbl}
\usepackage{pifont}
\definecolor{Gray}{gray}{0.9}

\ifcamera \usepackage[accsupp]{axessibility} \fi

% ...

\ifarxiv  \fi

\newcommand{\R}[1]{{%
    \textbf{%
        \ifstrequal{#1}{1}{\textcolor{red}{R#1}}{%
        \ifstrequal{#1}{2}{\textcolor{blue}{R#1}}{%
        \ifstrequal{#1}{3}{\textcolor{magenta}{R#1}}{%
        \ifstrequal{#1}{4}{\textcolor{teal}{R#1}}{%
                           \textcolor{cyan}{R#1}%
        }}}}%
    }%
}}

\usepackage{xr-hyper}

\makeatletter
\newcommand*{\addFileDependency}[1]{
  \typeout{(#1)}
  \@addtofilelist{#1}
  \IfFileExists{#1}{}{\typeout{No file #1.}}
}

\makeatother

\usepackage[pagebackref,breaklinks,colorlinks]{hyperref}
\usepackage[capitalize]{cleveref}
\crefname{section}{Sec.}{Secs.}
\crefname{table}{Table}{Tables}
\crefname{figure}{Fig.}{Figs.}

\begin{document}
%% TITLE
\title{\paperTitle}

\author{\authorBlock}
\maketitle

\begin{abstract}
The emergence of deepfake technologies has become a matter of social concern as they pose threats to individual privacy and public security. It is now of great significance to develop reliable deepfake detectors. However, with numerous face manipulation algorithms present, it is almost impossible to collect sufficient representative fake faces, and it is hard for existing detectors to generalize to all types of manipulation. Therefore, we turn to learn the distribution of real faces, and indirectly identify fake images that deviate from the real face distribution. In this study, we propose Real Face Foundation Representation Learning (RFFR), which aims to learn a general representation from large-scale real face datasets and detect potential artifacts outside the distribution of RFFR. Specifically, we train a model on real face datasets by masked image modeling (MIM), which results in a discrepancy between input faces and the reconstructed ones when applying the model on fake samples. This discrepancy reveals the low-level artifacts not contained in RFFR, making it easier to build a deepfake detector sensitive to all kinds of potential artifacts outside the distribution of RFFR. Extensive experiments demonstrate that our method brings about better generalization performance, as it significantly outperforms the state-of-the-art methods in cross-manipulation evaluations, and has the potential to further improve by introducing extra real faces for training RFFR. 

\end{abstract}
\section{Introduction}
\label{sec:intro}

Deepfake technologies~\cite{deepfakes} have become a growing concern of the society. Deepfake algorithms from recent studies~\cite{faceshifter, megapix, firstordermotion} are able to create increasingly realistic face images and videos, which can be maliciously used to spread misinformation. To tackle this issue, much effort has been put into the detection of deepfakes~\cite{ff, celeb-df, blink, headposes, va, forensictransfer}. While there exists some promising results in detecting particular manipulations~\cite{ff}, recent studies reveals that the performance of existing detection models drops drastically when presented with images manipulated with unseen methods~\cite{xray, celeb-df}. This deems generalization a major challenge for deepfake detection, as it is difficult to determine manipulation methods when suspicious images emerge in practice. 

\begin{figure}
\begin{center}
   \includegraphics[width=1\linewidth]{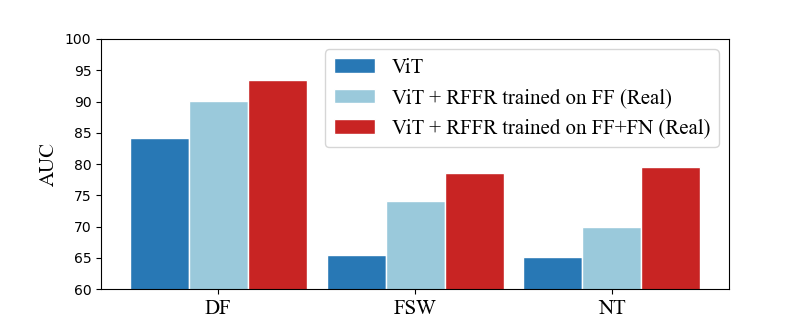}
\end{center}
	\vspace{-1.5em}
   \caption{The downstream task of deepfake detection consistently achieves better generalization performance with the use of RFFR, and this performance can be further improved by training RFFR with additional real face datasets.}
   \vspace{-1.5em}
\label{fig:extra_data}
\end{figure}

A number of studies in recent years devote to performing generalized deepfake detection~\cite{xray, lip, realforensics, f3net, gfsl, recce, multiatt, sola, sladd, sbi, dcl, uia} where a model is required to identify face images or videos manipulated with methods unseen during training. Though some progress are made, existing results on cross-manipulation tests are still far from satisfying. One major problem is that the models are generally trained on a limited source of fake data~\cite{ff, celeb-df, dfdc} and learn features that may not be adaptable to unseen manipulations. 

The inherent inaccessibility of sufficient fake face images limits the ability of models to identify fake images in all possible forms. Each deepfake algorithm leaves different traces on their creations, and the sheer number of existing algorithms denies the possibility of collecting enough representative types of manipulated images. However, real faces are considered to have a well-defined distribution without potentially unknown forms. We intend to focus on learning the distribution of the real face images, and treat fake faces as anomalies with rare occurrences~\cite{ADReview}. The priority of such a model is to gain sufficient knowledge of real faces, such as their structure and local texture, and be able to identify fake samples that deviate from the real face distribution. 

In this study, we propose Real Face Foundation Representation Learning (RFFR), which aims at learning the representation of real faces and detect the fake faces outside the distribution of RFFR. Specifically, we use a large number of real face images to train a model by masked image modeling (MIM)~\cite{simmim}. This model inpaints partially masked images with information from visible regions. By training exclusively on real faces, it is expected to implicitly learn the representation of real faces, or RFFR, and then reconstruct the masked regions based on its knowledge of the structure and texture of real faces. Applying this model to locally masked real images tends to result in a faithful reconstruction. On the other hand, since the model infers masked contents based on RFFR, it restores forged images into artifacts-free images with differences between its input and output. Such a discrepancy allows us to use a simple difference operation between the original masked image region and its recovered version to find a residual image, which signals the presence of potential anomalies in the region. 

Lastly, we combine residual and original image blocks as input, and train another dual-branch Vision Transformer~\cite{vit} as a deepfake detector. With the support of RFFR, our classifier demonstrates impressive abilities of generalizing to unseen fake images and avoiding severe overfitting during training. Additionally, we showcase a remarkable scalability of this framework, where representation learning aided by extra real faces consistently enhances downstream generalization, as illustrated in  ~\cref{fig:extra_data}.

We summarize our contribution as follows:

(1)	We propose to learn the distribution of real face images with Real Face Foundation Representation Learning (RFFR), which aims at improving downstream tasks with the learned representation. 

(2) We show the residual images obtained by inferring with our RFFR model highlights potential artifacts left by face manipulation algorithms, and therefore facilitates the identification of deepfake images.

(3) Extensive experiments on cross-manipulation and cross-dataset benchmarks for deepfake detection show that RFFR brings about a scalable generalization performance as well as remarkable resistance to overfitting, and significantly outperforms the state-of-the-art methods.

\section{Related Work}
\label{sec:related}

We briefly review related works of deepfake detection, representation learning and anomaly detection.

\subsection{Deepfake Detection}
\textbf{Traditional methods.} 
Deepfake detection has drawn much attention since the first emergence of face manipulation algorithms~\cite{deepfakes}. Traditional methods use pre-determined features to spot the imperfections in deepfakes, such as inconsistent headposes~\cite{headposes} and eye-blinking~\cite{blink}. On the other hand, the introduction of large deepfakes datasets, such as Faceforensics++~\cite{ff}, fuels the development of learning-based approaches. With the help of such datasets, it is found that without designating specific features, a deep CNN-based binary classifier performs well enough on recognizing specific manipulations, given the model has been trained on images created with this particular manipulation~\cite{ff}.

\textbf{Cross-manipulation methods.} As a result of drastic differences between various types of manipulations, it is hard for a model to detect the artifacts of manipulations that are not contained in training set~\cite{xray, celeb-df}. Therefore, many works turn to focus on improving the generalization performance of deepfake detectors, most of which are based on the extensions of earlier feature-based methods. They utilize the most common artifacts available in existing datasets, such as inconsistency in videos~\cite{lip}, blending boundaries~\cite{xray}, biological signals~\cite{fake_catcher}, and frequency artifacts~\cite{f3net, SPSL}. Although these features are proven effective, many rely on glitches made by primitive manipulations, which may not be detectable in further improved manipulations. To facilitate generalization to a broader range of manipulations, some studies enrich the distribution of fake data by creating new fake face images~\cite{sladd, sbi}, and bring significant improvements over existing methods. 

\textbf{Reconstruction learning-based methods.} Reconstruction learning for deepfake detection has emerged in some pioneering works. One method~\cite{ocfakedect} trains autoencoders with real samples and directly identify samples with large reconstruction error as deepfakes. Denoising~\cite{recce}, colorization, and super-resolution~\cite{beyondspectrum} have been used as reconstruction targets for real samples as well. Compared to these whole-image reconstruction approaches, our method uses masked image modeling to learn real face representations more effectively and yields better generalization performance.

\subsection{Representation Learning with Unlabeled Data}
Computer vision tasks have greatly benefited from pre-training on large datasets. In recent years, self-supervised pre-training is gaining increasing popularity as they learn effective representations without any labeled data. Earlier methods learn representations with pretext tasks like solving jigsaws~\cite{jigsaw}, predicting rotations~\cite{rotation}, and colorization~\cite{colorization}. More recently, contrastive learning~\cite{simclr, moco} learns to aggregate different views of the same image and disperse different images in the representation space. Masked image modeling (MIM)~\cite{beit, mae, simmim} trains models to predict masked regions of input images, a training mechanism we adopt for RFFR. By learning representations with effective pretext tasks, models for specific vision tasks can achieve significantly better results than learning from scratch.

Besides learning general image representations, learning with data from certain domains can also facilitate specific downstream tasks~\cite{realforensics, voice-face, faceunit}. Among this line of work, RealForensics~\cite{realforensics} is the most similar work to ours. This study uses contrastive learning to learn audiovisual representations of real face videos in the hope of improving deepfake detection. However, learning from static real face images, the direct sources of face manipulations, is still left unexplored. In this work, we use MIM-based representation learning to show that without temporal features or audio correspondence, learning the representation of static faces provides abundant information to facilitate deepfake detection as well.

\begin{figure*}
\begin{center}
\vspace{-1.5em}
   \includegraphics[width=0.95\linewidth]{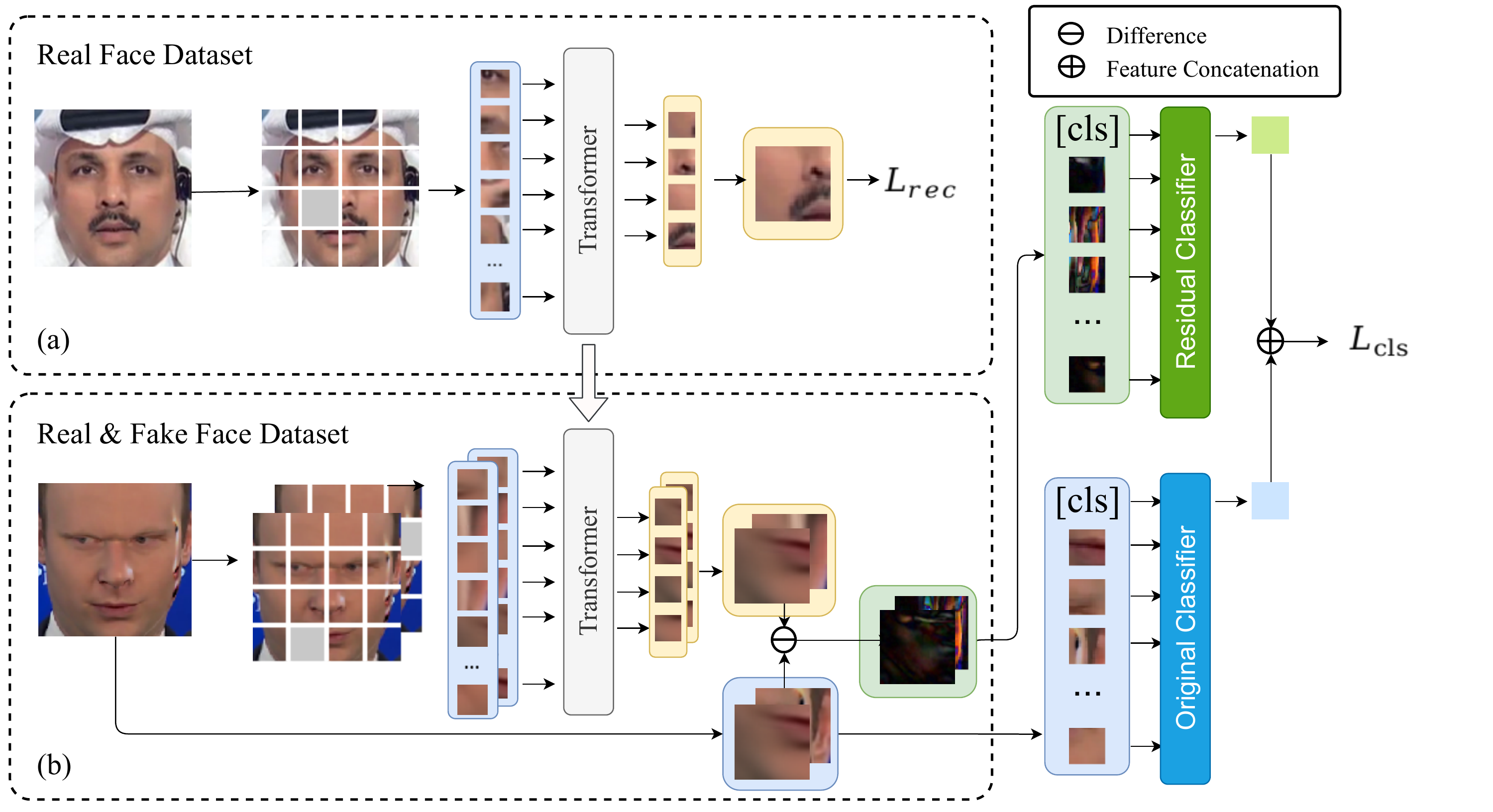}
\end{center}
\vspace{-2em}
   \caption{Pipeline of our RFFR-based deepfake detection. (a) We use real faces to train a representation learning model by masked image modeling, which learns to reconstruct masked regions of real faces by minimizing reconstruction error $L_{rec}$. We expect this model to encode masked faces to real face foundation representations (RFFR) and decode the RFFRs to faithfully reconstruct masked real images, but fail at fake images. 
   (b) We mask random blocks of a suspected image for the trained inpainter to reconstruct. The original image blocks are subtracted from the output of the model to create residual image blocks, which signals artifacts. We train a dual-branch ViT classifier with $L_{cls}$ to identify deepfakes with both the original image blocks and the residual image blocks. }
\label{fig:pipeline}
\vspace{-1.5em}
\end{figure*}

\subsection{Anomaly Detection}
Anomaly detection is another task closely related to our research, which aims to identify abnormal data that deviate from a specific distribution. One strategy of anomaly detection is implicitly learning the representation of normality by reconstructing normal samples. Anomalies are identified as they can not be properly reconstructed.  ~\cite{ADReview}.

A major problem of this approach is that autoencoders often generalize well enough for both normal and abnormal samples. To avoid learning such identity mappings, some studies equip the autoencoder with a memory~\cite{memae, block_memory} to perform reconstruction based on only limited prototypes. Others~\cite{ocgan} introduce discriminators to adversarially learn distributions that strictly corresponds to the normal data distribution. In this work, we turn to train MIM models, so that local image blocks are predicted based on the neighborhood instead of themselves, thus avoiding identity mappings.

\setlength{\abovedisplayskip}{5pt} % 对公式上下间隔有用
\setlength{\belowdisplayskip}{5pt}

\section{Method}
\label{sec:method}

\subsection{Overview}

Deepfake detectors suffer from severe performance drops when applied to unseen manipulations, because artifacts of specific manipulations are hardly generalizable features. At the same time,  it is almost impossible to collect sufficient representative fake images to ensure model awareness of all possible manipulations, considering the diversity and ever-developing nature of deepfakes. Therefore, instead of focusing on the fake faces, we set out to learn an accurate real face distribution, and aim at identifying fake faces that violate the boundary of this distribution. In this work, we propose Real Face Foundation Representation Learning (RFFR) to achieve this purpose. We attempt to learn rich representations from large-scale real face datasets, so the model is able to identify fake faces outside the distribution of RFFR.

As shown in \cref{fig:pipeline}, we use real faces to train a model by masked image modeling (MIM), which masks a random block from an input image and learns to recover it based on the rest of the image. The trained model inpaints masked real faces well, but it tends to fail at inpainting fake faces based on the representations learned with real faces. This creates a discrepancy between the input and output in the form of a residual image block. We train a classifier with residual image blocks and corresponding original image blocks to perform deepfake detection. 

\subsection{Real Face Foundation Representation Learning}

\begin{figure*}
\vspace{-2em}
\begin{center}
   \includegraphics[width=1\linewidth]{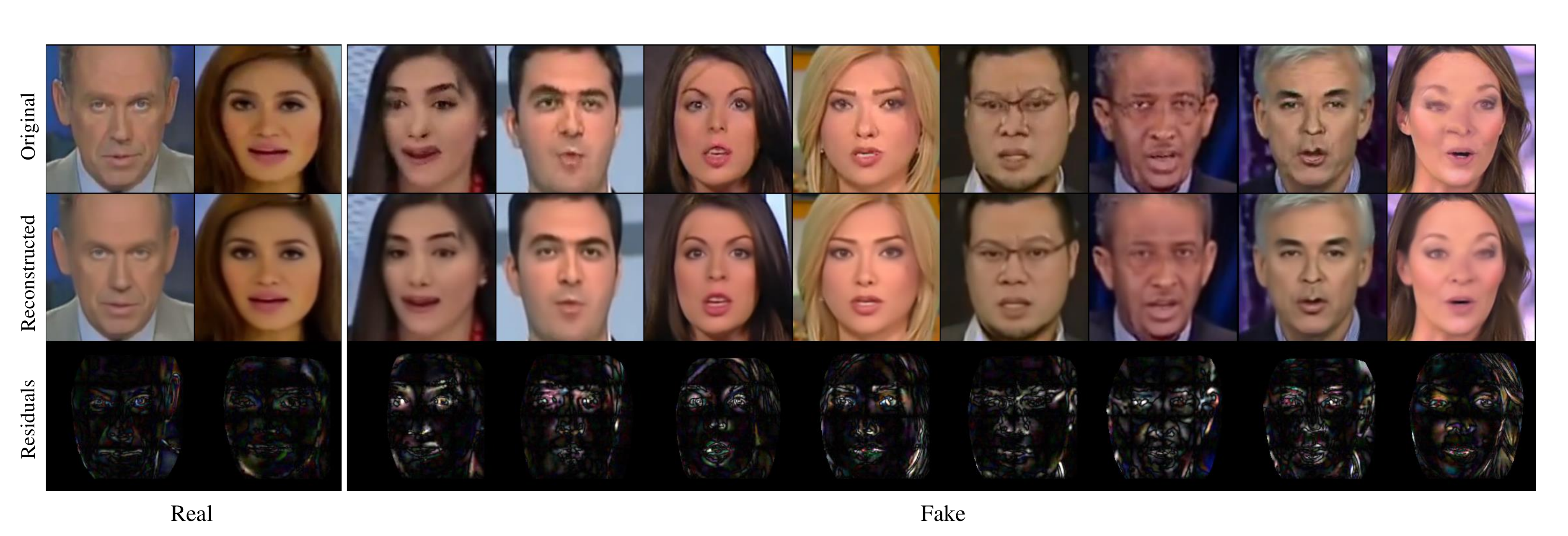}
\end{center}
\vspace{-2em}
   \caption{Visualizations of processing real and fake samples with our RFFR-learning inpainting model. Clear artifacts in the original fake images (first row) are effectively removed in the reconstructed images (second row) and highlighed in the residual images (third row). Owing to an inevitable information loss during inpainting, the real residuals on the left are not completely empty, but they are in contrast with the fake ones that indicate clear forgery patterns.}
\vspace{-1.5em}
\label{fig:residual}
\end{figure*}

We leverage MIM to train a representation learning model with real faces to learn the RFFR. Given a locally masked real face image, the model learns to reconstruct the masked region based on the visible region of this image. Specifically, given a real face image $X^i$,  we divide it into a grid of $k\times k$ image blocks with a division rule $g$, where each block is to be individually masked and restored later:
\begin{equation}
  B_1^i, B_2^i, \cdots, B_{k\times k}^i  = g(X^i).
  \label{eq:divide}
\end{equation}
The MIM model $M$ accepts an input image $X^i$ along with a randomly selected mask $m_j$, and reconstructs the masked block to obtain $\hat{B^i}$:
\begin{equation}
  \hat{B^i} = M(X^i, m_j),
  \label{eq:inpaint}
\end{equation}
where we ensure the $j$-th mask covers the $j$-th block of an image. For coordinate $(x, y)$,
\begin{equation}
  m_j (x, y) = \left\{
  \begin{aligned}
  0, \ \ \ &(x, y) \in B_j	\\
  1, \ \ \ &otherwise	\\
  \end{aligned}
  \right..
  \label{eq:mask}
\end{equation}Finally, this representation model is trained with a simple $L_2$ loss
\begin{equation}
  L_{rep} = \sum_{i=1}^n||\hat{B^i_j} - B^i_j||_2,
  \label{eq:l2}
\end{equation}
so that the model learns to inpaint masked images with supervisory signal from the images themselves.

By learning to inpaint any random region of real face images, the model is expected to comprehensively learn real face representations, or RFFR. When provided with a masked real face image, the model infers the representation of the face and decodes this representation by generating a reasonable image block to fill in the masked region. Upon processing fake samples, this model tends to infer a real face representation based on its input and use real textures to restore masked regions of the samples. This creates a large discrepancy between the input and output of the model, which signals low-level artifacts present in the fake images.

In \cref{fig:residual}, we visualize both real and fake samples processed by our RFFR-learning inpainting model. For each image, we iteratively mask all $k\times k$ blocks, restore them with the inpainting model, and collect all output blocks to assemble a whole reconstructed image. We subsequently subtract the original image from the reconstructed image to obtain the residual image (amplified for better visibility). The reconstructed faces closely resemble the original faces with successful reconstructions of their high-level facial attributes. This preservation of high-level semantic information benefits from the learning of RFFR. However, low-level artifacts in the fake images, many clear enough to be observable on the foreheads, eyes, mouths, \etc, are effectively removed, as they are not represented in the distribution of RFFR. This also results in highlights in the residual images, which provides essential information to guide deepfake detection.

\subsection{Deepfake Detection}
\label{sec:method_deepfake_detection}
The previous section guarantees that we have a compact distribution of real faces, which can be used to reveal low-level artifacts outside the distribution of RFFR. This revelation is instantiated by the residual images we describe above. In this section, we detail the process of leveraging residual images generated with RFFR to train a deepfake detector. 

Formally, we obtain residual image blocks by subtracting the original block from the block generated with RFFR:
\begin{equation}
  R_j^i = \alpha(\hat{B_j^i} - B_j^i),
  \label{eq:difference}
\end{equation}
where we use a constant factor $\alpha$ to amplify the subtle signals in the residuals and match its scale to that of natural images.

Unlike visualizations in \cref{fig:residual}, we do not generate the whole residual image for training the classifier. Instead, we develop a random input mechanism, which randomly selects a subset of all image blocks to enter the classifier. Every block in an image is selected with a pre-determined probability $p$. Upon selecting block $B_j^i$ for input, we invoke \cref{eq:inpaint} and \cref{eq:difference} to generate corresponding residual image block $R_j^i$. This process is repeated until we make the decision for each block whether to utilize it for classification or not. Eventually, we obtain residual image blocks $ \{R_j^i\}_{j = k_1, k_2, \cdots}$ and their corresponding original image blocks $\{B_j^i\}_{j = k_1, k_2, \cdots}$, where $k_1, k_2, \cdots$ are the indexes of all selected blocks. 

This random input mechanism benefits our deepfake detector in two ways. Firstly, compared to random input, a complete reconstruction carried out by an inpainting model is very time-consuming. To obtain residual images in the form of~\cref{fig:residual}, we need a total of $k\times k$ inferences for every batch of images. This seriously prolongs the training process. Secondly, random input improves generalization, as we show in the ablation study. We hypothesize that learning with randomly selected blocks reduces overfitting in that the model is forced to learn from different locations. A model that accepts full images tends to focus on the most suspicious regions of artifacts. By only providing a subset of all image blocks to the classifier, we, in effect, mask out the rest of the blocks. The model then learns with artifacts from random locations, which could be too subtle to be spotted by the model when prominent artifacts are within the input image~\cite{rfm, gfsl}. This allows the model to form a complete set of feature revealed with RFFR, thus improving generalization.

With both sets of image blocks collected, we integrate them as input to enter a classifier $F$ to perform deepfake detection. Two Vision Transformers (ViTs)~\cite{vit} are adopted to form a dual-branch classifier, with each accepting one set of image blocks. They jointly produce one prediction $\hat{Y^i}$:

\begin{equation}
  \hat{Y^i} = F(R_{k_1}^i, R_{k_2}^i, \cdots ; B_{k_1}^i, B_{k_2}^i, \cdots).
  \label{eq:detect}
\end{equation}
As we use ViTs for training, the blocks are further broken down into smaller patches to enter the network. In addition, each patch is aided with its own position embedding, which helps the model better identify artifacts in images based on their specific locations. At the end of processing, each ViT branch generates a class token, and the two tokens are merged to create a final feature for the current input image and subsequently a prediction. 

Finally, given the ground truth $Y^i$s, we train the deepfake detector with a simple classification loss:

\begin{equation}
  L_{cls} = -\sum_{i=1}^n Y^i log(\hat{Y^i})
  \label{eq:ce}
\end{equation}

\section{Experiments}
\label{sec:experiments}

\subsection{Setup}
\textbf{Datasets.} We evaluate RFFR with four challenging datasets specifically designed for deepfake detection. We adopt the high quality (HQ) version of Faceforensics++ (FF)~\cite{ff} for training our deepfake detector. Faceforensics++ includes videos of real faces as well as four subsets of fake faces, each manipulated with a different algorithm, namely Deepfakes (DF), Face2Face (F2F), FaceSwap (FSW) and NeuralTextures (NT). We also utilize the test set of Celeb-DF~\cite{celeb-df} and DFDC~\cite{dfdc} for evaluating the cross-dataset performance of our model. Finally, in addition to real faces of Faceforensics++, we adopt the real face images from ForgeryNet (FN)~\cite{forgerynet} for learning RFFR, which helps improve representation learning with additional data.

\textbf{Implementation Details.} We extract the frames from all video datasets and use RetinaFace~\cite{retinaface} to detect and align the faces. All images are scaled to the size of $224 \times 224$. For our RFFR model, we adopt a base version of Masked Autoencoder (MAE)~\cite{mae} and train it on real faces with a batch size of $128$. Following MAE, we set the learning rate at $7.5 \times 10^{-5}$ and adjust it with a schedule with warmup and cosine decay. By default, we train this model with the real faces from both FF~\cite{ff} and FN~\cite{forgerynet}. 

For training the deepfake detector, we divide each image with $k = 4$ (Refer to Appendix for the motivation of choosing $k$). Each block enters the classifier with a probability of $p = 0.25$, and the residual images are amplified by $\alpha=4$. No data augmentation is applied to the images. We initialize both branches of Vision Transformer with ImageNet-pretrained weights and train them with a learning rate of $2 \times 10^{-5}$. During testing, we iteratively mask and restore all blocks to obtain a full residual image for the detector to process. We evaluate the testing results with AUC (Area Under Curve). 

\subsection{Cross-domain performance evaluation}
In this section, we test the performance of our RFFR-based deepfake detector with cross-manipulation and cross-dataset evaluations. 

\textbf{Cross-manipulation evaluations.} We train our deepfake detector on each subset of Faceforensics++ and test on all four subsets to demonstrate our model's ability to identify different manipulations, including those not seen during training. \emph{We adopt the HQ version of FF for both training and testing, and only use one frame every video for testing.} We compare our results with state-of-the-art image-based methods Multi-Attention~\cite{multiatt}, DCL~\cite{dcl}, RECCE~\cite{recce} and UIA-ViT~\cite{uia}. We ran the public code of RECCE and UIA-ViT to produce results under the same setting.

In~\cref{tab:cross-manipulation}, we show that our method outperforms the state-of-the-art methods under most settings, with a maximum improvement of $10.25\%$ (F2F $\rightarrow$FSW). Meanwhile, our model remains effective under the four intra-domain settings, which are shown in gray. The method tends to slightly underperform when trained on NeuralTextures, likely because its manipulation patterns only exist in certain small regions, and may be neglected during our block sampling. Nevertheless, compared to existing methods, our deepfake detector yields much better overall performances. 

\begin{table}[t]
\setlength\tabcolsep{4.5pt} 
\caption{Cross-manipulation performances in terms of AUC(\%) compared with previous methods. Classifiers are trained on one subset of FF and tested on all four subsets. Intra-domain results are marked in gray. We ran the public code of methods marked with "*" to produce results under identical settings \emph{(HQ for training and single frames for testing).}}
\vspace{-1.5em}
\label{tab:cross-manipulation}
\begin{center}  
\scalebox{0.80}{
\begin{tabular}{c|l|cccc|c}
\toprule
Training &\multirow{2}*{Method} & \multicolumn{4}{c|}{Test data} & \multirow{2}*{Avg} \\
\cmidrule(lr){3-6}
     data  &            ~                   & DF    & F2F   & FSW   & NT    & ~   \\
     
\midrule
\multirow{5}*{DF}
& MultiAtt~\cite{multiatt} & \cellcolor{Gray}99.92 & 75.23 & 40.61 & 71.08 & 71.71                \\ 
& DCL~\cite{dcl}       & \cellcolor{Gray}\textbf{99.98} & \textbf{77.13} & 61.01 & 75.01 & 78.28              \\
& RECCE*~\cite{recce}     & \cellcolor{Gray}99.19 & 74.39 & 57.42 & \textbf{85.04} & 79.01                \\ 
& UIA-ViT*~\cite{uia}  & \cellcolor{Gray}99.39      &   74.44    &   53.89    &   70.92    & 74.66 \\ 
& Ours  & \cellcolor{Gray}99.19 & 76.61 & \textbf{68.96} & 74.83 & \textbf{79.90}            \\ 
       
\midrule
\multirow{5}*{F2F}
        & MultiAtt~\cite{multiatt}       & 86.15 & \cellcolor{Gray}99.13 & 60.14 & 64.59 & 77.50 \\
        & DCL~\cite{dcl}       & 91.91 & \cellcolor{Gray}99.21 & 59.58 & 66.67 & 79.34 \\
       & RECCE*~\cite{recce}       & 88.04 & \cellcolor{Gray}98.93 & 67.35 & 74.16 & 82.12 \\
       & UIA-ViT*~\cite{uia}       & 83.39 & \cellcolor{Gray}98.32 & 68.37 & 67.17 & 79.31 \\
       & Ours                                  & \textbf{93.75} & \cellcolor{Gray}\textbf{99.61} & \textbf{78.62} & \textbf{79.56} & \textbf{87.81} \\

\midrule
\multirow{5}*{FSW}
& MultiAtt~\cite{multiatt} & 64.13 & 66.39 & \cellcolor{Gray}99.67 & 50.10 & 70.07              \\
& DCL~\cite{dcl}           & 74.80 & 69.75 & \cellcolor{Gray}99.90 & 52.60 & 74.26              \\
& RECCE*~\cite{recce}       & 66.66 & 73.66 & \cellcolor{Gray}\textbf{99.76} & \textbf{57.46} & 74.39               \\

& UIA-ViT*~\cite{uia}       &   81.02    &   66.30    & \cellcolor{Gray}99.04      &   49.26    & 73.91 \\ 
& Ours                                           & \textbf{87.46} & \textbf{75.96} & \cellcolor{Gray}99.42 & 55.87 & \textbf{79.68}            \\ 

\midrule
\multirow{5}*{NT}
& MultiAtt~\cite{multiatt} & 87.23 & 75.33 & 48.22 & \cellcolor{Gray}98.66 & 77.36                \\
& DCL~\cite{dcl}      & 91.23 & 79.31 & 52.13 & \cellcolor{Gray}\textbf{98.97} & 80.41                \\
& RECCE*~\cite{recce}    & \textbf{90.20}  & 76.65 & \textbf{58.06} & \cellcolor{Gray}97.17 & \textbf{80.52}                \\
 & UIA-ViT*~\cite{uia}  &    79.37   &   67.98    &   45.94    &\cellcolor{Gray}94.59       & 71.97 \\
 & Ours     & 84.31 & \textbf{81.04} & 54.67 & \cellcolor{Gray}96.19 & 79.05          \\
       
\bottomrule
\end{tabular}}
\vspace{-2em}
\end{center}
\end{table}

\textbf{Cross-dataset evaluations.} We train our model on the Faceforensics++ dataset and evaluate its performance on the test sets of Celeb-DF\cite{celeb-df} and DFDC~\cite{dfdc}. Specifically, following the previous practice in~\cite{lip}, we validate the model on Celeb-DF and use the selected model to test on DFDC.  \emph{We adopt the HQ version of FF for training, and only use one frame every video for testing.} Under the same setting, we ran the public code of RECCE~\cite{recce}, UIA-ViT~\cite{uia} and SBI~\cite{sbi} to produce corresponding results. In Table~\ref{tab:cross-dataset}, we show a competitive performance with existing image-based methods, signaling satisfying adaptability of RFFR to different datasets, especially high quality datasets like Celeb-DF. 
  
SBI~\cite{sbi} is a recent powerful deepfake detection method. By utilizing a hand-crafted blending algorithm to produce diverse fake samples, it achieves highly competitive performances on datasets including Celeb-DF. We show that by training on fake samples generated by SBI, our approach can further improve upon their state-of-the-art result. 

\begin{table}[]
\setlength\tabcolsep{4.5pt} 
\caption{Cross-dataset performances in terms of AUC(\%) compared with previous methods. Classifiers are trained on FF and tested on Celeb-DF and DFDC. We ran the public code of methods marked with "*" to produce results under identical settings \emph{(HQ for training and single frames for testing).}}
\vspace{-1em}
\label{tab:cross-dataset}
\begin{center}  
\scalebox{0.90}{
\begin{tabular}{l|cc}
\toprule
\multirow{2}*{Method} & \multicolumn{2}{c}{Test data}\\
\cmidrule{2-3}
        ~                           &     Celeb-DF         &  DFDC \\
\midrule
      Xception~\cite{xception}  &     65.30       &    -  \\
      Face X-ray~\cite{xray}          &     74.20       &     70.00 \\
      MultiAtt~\cite{multiatt}        &     67.44       &     67.34 \\
      SPSL~\cite{SPSL}                &     76.88        &   -  \\
      SOLA~\cite{sola}                &       76.02         &  -    \\
      SLADD~\cite{sladd}              &    79.70       &  -  \\
      RECCE*~\cite{recce}             &     68.94       &   68.34   \\
      UIA-ViT*~\cite{uia}             &     80.31      &   67.93   \\
      SBI*~\cite{sbi}                       &       86.46     &   66.60     \\
\midrule
 	Ours                                      &   81.97  & \textbf{72.08}  \\
    Ours + SBI~\cite{sbi}                  &  \textbf{88.98}           &    67.84   \\
\bottomrule
\end{tabular}}
\vspace{-2.5em}
\end{center}
\end{table}

\subsection{Ablation Study}
\label{ablation}

In this section, we analyze the effect of our implementations for RFFR learning and deepfake detection. 

\textbf{Effect of the training data for RFFR.} The effectiveness of deepfake detection with RFFR depends on the quality of representation learning, where the real faces plays an important role. In this experiment, we examine the effect of scaling the real face dataset for representation learning. As a baseline, we learn RFFR with only real faces from Faceforensics++ (FF), the same data we use for the downstream classification tasks. Meanwhile, another model is supplemented with real faces from both FF and ForgeryNet (FN), a significantly larger and more diverse dataset. We train deepfake detectors on the F2F subset of FF with residual images produced by these two models. In Table~\ref{tab:data}, we demonstrate that including the extra dataset of ForgeryNet for learning RFFR consistently improves the performances of the deepfake detector in all tests, creating a maximum performance gain of $9.57\%$  in terms of AUC (F2F $\rightarrow$ NT).

We note that learning RFFR with FF already allows our deepfake detector to outperform the state-of-the-arts. Nevertheless, learning with extra data enhances the efficacy of our real face foundation representations, and further improves the downstream task of deepfake detection. Therefore, refining the representation learning of real faces, especially with large-scale datasets, could be a viable path for further improving generalized deepfake detection. 

\begin{figure}
\centering
  \includegraphics[width=0.9\columnwidth]{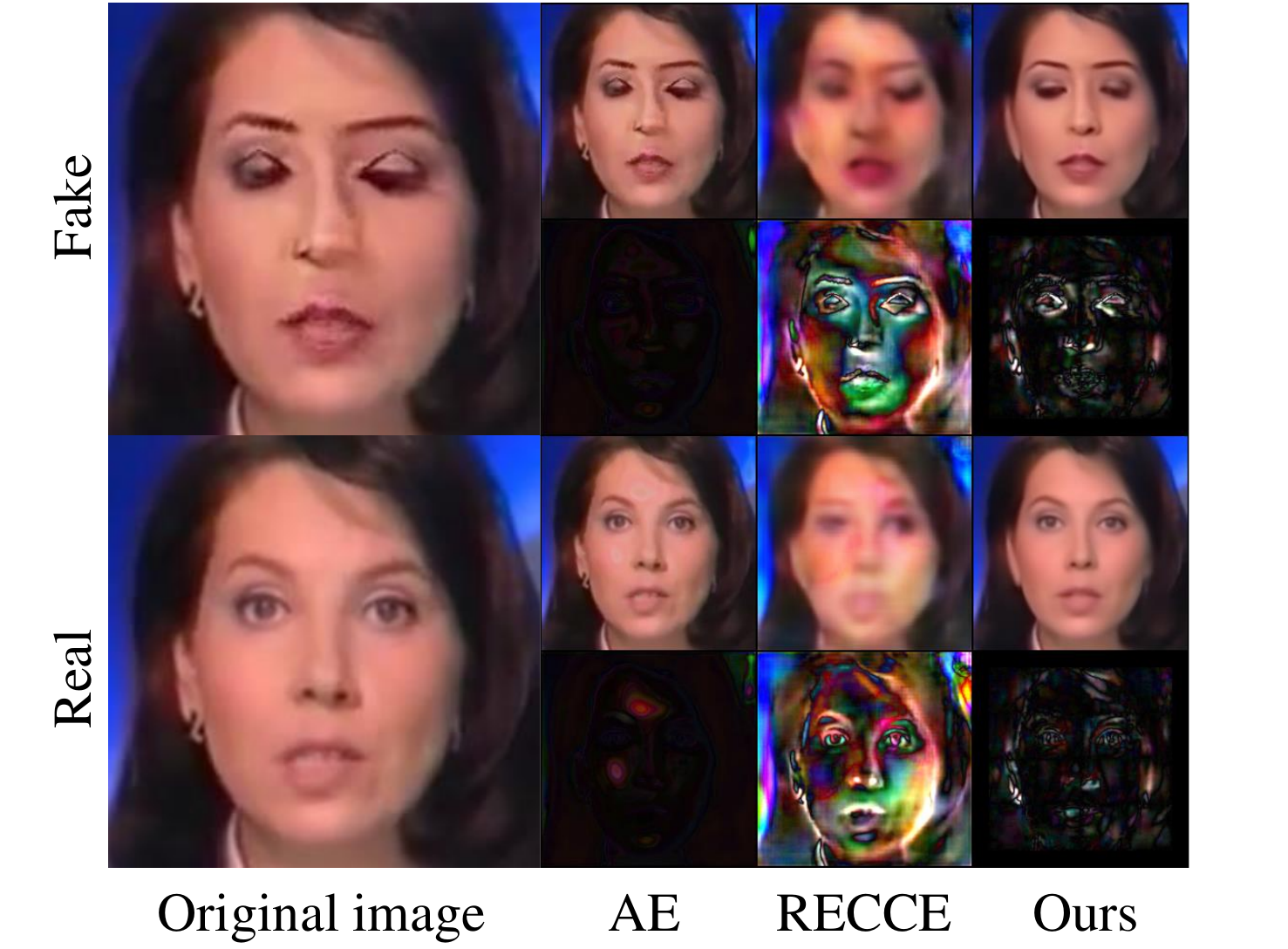}
  \vspace{-1em}
   \caption{Reconstruction results and residual images of the autoencoder (AE), RECCE~\cite{recce} and our inpainting model. AE reconstructs both images perfectly, leaving no information in residual images. RECCE~\cite{recce} suffers from insufficient training. Our model successfully highlights potential artifacts in the residual image of only the fake face, and therefore can best facilitate deepfake detection. }
\vspace{-1em}
\label{fig:unet_comparison}
\end{figure}

In addition, we examine the scalability of RECCE under the same setting, considering that RECCE~\cite{recce} also involves learning to reconstruct real samples for deepfake detection. However, their performance gain is less significant than ours. Although the reconstruction branch of RECCE~\cite{recce} is able to highlight forgery cues with residual images, they tend to involve more background noise caused by imperfect reconstructions, as depicted in~\cref{fig:unet_comparison},. This undermines the ability of residual images to expose artifacts for deepfake detection. 

\begin{table}[t]
\setlength\tabcolsep{4.5pt} 
\caption{Deepfake detection performances of RECCE~\cite{recce} and our method with different real face dataset, namely the real faces from Faceforensics++ (FF) alone, and FF combined with ForgeryNet (FF + FN). Classifiers are trained on F2F and tested on four subsets of FF. We present the results in AUC (\%).  }
\vspace{-1.5em}
\label{tab:data}
\begin{center}  
\scalebox{0.90}{
\begin{tabular}{c|c|cccc|c}
\toprule
\multirow{2}*{Method} & Real face  & \multicolumn{4}{c|}{Test data} & \multirow{2}*{Avg} \\
\cmidrule(lr){3-6}
&dataset  &      DF    & F2F   & FSW   & NT    & ~   \\
    \midrule
\multirow{2}*{RECCE~\cite{recce}}&FF           & 88.04          & 98.93          & 67.35          & 74.16          & 82.12          \\
&FN + FF &  90.12       & 99.24       & 69.89    & 79.59     & 84.71		\\
    \midrule
\multirow{2}*{Ours}&FF           & 90.16          & 98.56          & 74.10          & 69.99          & 83.20          \\
&FN + FF & \textbf{93.44}       & \textbf{99.61}        & \textbf{78.62}       & \textbf{79.56}        & \textbf{87.81}		\\
\bottomrule
\end{tabular}}
\vspace{-1em}
\end{center}
\end{table}

\textbf{Effect of masked image modeling for RFFR.} We analyze the effect of using MIM-based residual images for deepfake detection. We train a UNet-based autoencoder (AE) to learn the reconstruction of real faces and obtain residual images. Our MIM-trained inpainting model and the AE are compared on the quality of reconstruction in~\cref{fig:unet_comparison}. Note that despite being trained with real faces, the AE "generalizes" well to fake images, preserving delicate details, including the artifacts caused by manipulations. Such generalization leaves the residual images empty with little information. 

Masked image modeling enables our model to learn better real face representations and inpaint fake faces with real textures instead of artifacts. In the downstream task of deepfake detection,  our classifier generalizes significantly better than the AE-based classifier, which performs only marginally better than learning with no residuals (detailed in Appendix). Both the reconstruction results and the downstream performance confirm the validity of our choice to learn RFFR with MIM instead of direct reconstruction.

\textbf{Effect of classifier backbone.} In Table~\ref{tab:backbone}, we present the deepfake detection results of vanilla Xception~\cite{xception} and Vision Transformer (ViT)~\cite{vit}, both trained with full original images. The models are trained with the F2F subset of FF and tested on all four subsets. While a larger backbone increases a deepfake detector's generalization performance in some cases, it is not the primary factor of our performance improvement. Instead, it is the residual input aided by RFFR that leads the performance gain.

\begin{table}[t]
\setlength\tabcolsep{4.5pt} 
\caption{Comparing ours results with vanilla backbones. We present the results in AUC (\%).  }
\label{tab:backbone}
\vspace{-1.5em}
\begin{center}  
\scalebox{0.90}{
\begin{tabular}{c|c|cccc|c}
\toprule
Training  &  \multirow{2}*{Method}    &   \multicolumn{4}{c|}{Test Data} & \multirow{2}*{Avg} \\
\cmidrule(lr){3-6}
 data  &   ~  &   DF    & F2F   & FSW   & NT    & ~   \\
    \midrule
\multirow{3}*{F2F} & Xception~\cite{xception} & 84.94          & 99.26          & 58.82          & 71.19          & 78.55          \\
                                   & ViT~\cite{vit}      & 84.25          & 97.89          & 65.53          & 65.18          & 78.21          \\
                                   & Ours     & \textbf{93.44} & \textbf{99.61} & \textbf{78.62} & \textbf{79.56} & \textbf{87.81} \\
\bottomrule
\end{tabular}}
\vspace{-1.5em}
\end{center}
\end{table}

\textbf{Effect of classifier design.} We compare different variants of our classifier design. Specifically, we analyze the performance gains brought by the introduction of two branches and the random input mechanism. We test six variants of our classifier by training them with the F2F subset of FF and testing with the FSW subset. The settings of these variants are specified by the input data they accept, as shown in~\cref{tab:classifier}. 

\begin{table}[t]
\caption{Deepfake detection performances with classifiers of different inputs in terms of AUC (\%). We train the classifiers on F2F and test on FSW.}
\label{tab:classifier}
\vspace{-1.5em}
\begin{center}
\begin{tabular}{c|c|c|c|c}
\toprule
\multicolumn{2}{c|}{Original Image} & \multicolumn{2}{c|}{Residual Image} & \multirow{2}*{AUC (\%)} \\
\cline{1-4}
               Full        &             Random           &          Full          &          Random          &   ~\\
 \hline
\checkmark        &                                       &                            &                                   &  65.53\\
% \hline
                              &                                      &   \checkmark    &                                   &  66.30  \\
 %\hline
\checkmark        &                                      &   \checkmark    &                                   &  71.48  \\
 %\hline
                             &       \checkmark          &                             &                                   &  70.76  \\
%\hline
                             &                                       &                             &      \checkmark       &  68.10  \\
 %\hline
                             &        \checkmark         &                             &      \checkmark       &  \textbf{78.62}  \\
\bottomrule
\end{tabular}
\vspace{-2em}
\end{center}
\end{table}

\begin{table*}[t]
\setlength\tabcolsep{4.5pt} 
\caption{Deepfake detection performances of validated and non-validated models. Classifiers are trained on F2F and tested on four subsets of FF. We present the results and the performance gaps in AUC (\%). Second best results are underlined. }
\label{tab:validation}
\vspace{-1em}
\begin{center}  
\scalebox{0.90}{
\begin{tabular}{c|c|llll|l}
\toprule
\multirow{2}*{Method}  & \multirow{2}*{Validated} & \multicolumn{4}{c|}{Test Data} & \multirow{2}*{Avg} \\
\cmidrule(lr){3-6}
~                   &                      ~                   &      DF               & F2F                    & FSW                 & NT                    & ~   \\
    \midrule
\multirow{2}*{Xception\cite{xception}} &   \checkmark    & 84.94                 & 99.26                & 58.82                 & 71.19                & 78.55            \\
~ &                                             -                              & 83.08   (- 1.86) & 99.12   (- 0.14) & 46.63   (- 12.19) & 64.93   (- 6.26)  & 73.44   (- 5.11)  \\
 \hline
 \multirow{2}*{RECCE\cite{recce}} &\checkmark               & 88.04                & 98.93                 & 67.35                & 74.16                & 82.12            \\
 ~&                                                -                  & 74.51   (- 8.57) & 99.22   (+ 0.29)  & 50.17   (- 17.18) & 59.46   (- 14.70)  & 70.84   (- 11.28) \\
 \hline
\multirow{2}*{Ours} &    \checkmark  & \textbf{93.44}            & \textbf{99.61}            & \textbf{78.62}            & \textbf{79.56}            & \textbf{87.81}            \\
 ~&  - & \underline{91.56} (- 1.88) & \underline{99.39}   (- 0.22) & \underline{76.00}   (- 2.62)  & \underline{76.41} (- 3.15) & \underline{85.84}   ( - 1.97)    \\
\hline
\end{tabular}}
\vspace{-2em}
\end{center}
\end{table*}

We treat the vanilla ViT with full original image input as a baseline, which achieves an AUC of $65.53\%$. By switching to accept the full residual images, we obtain a $0.77\%$ performance gain. Combining the two modalities to form a dual-branch classifier further increases our result to $71.48\%$. This demonstrates that the artifacts are better exploited when both the original and the residual images enter the classifier, and are used as references to each other. Therefore, both modalities should be considered for classification. 

In addition, we improve on the test by merely modifying the baseline ViT to accept randomly selected original image blocks. This results in a $5.23\%$ increase in performance. Similarly, changing full residual input to random residual blocks also results in a $1.8\%$ improvement. These observations confirm our hypothesis in \cref{sec:method_deepfake_detection} that models benefit from learning with random inputs, which prevents the model from only focusing on the most prominent features in an image, and forces it to learn from subtle artifacts. 

Finally, bringing in the random input mechanism for the dual-branch classifier completes our full implementation, which maximally exploits the artifacts exposed by RFFR and achieves the best performance of $78.62\%$.

\subsection{Validation-free Model Selection}
\label{sec:validation-free}

\begin{figure}
\centering
  \includegraphics[width=0.5\textwidth]{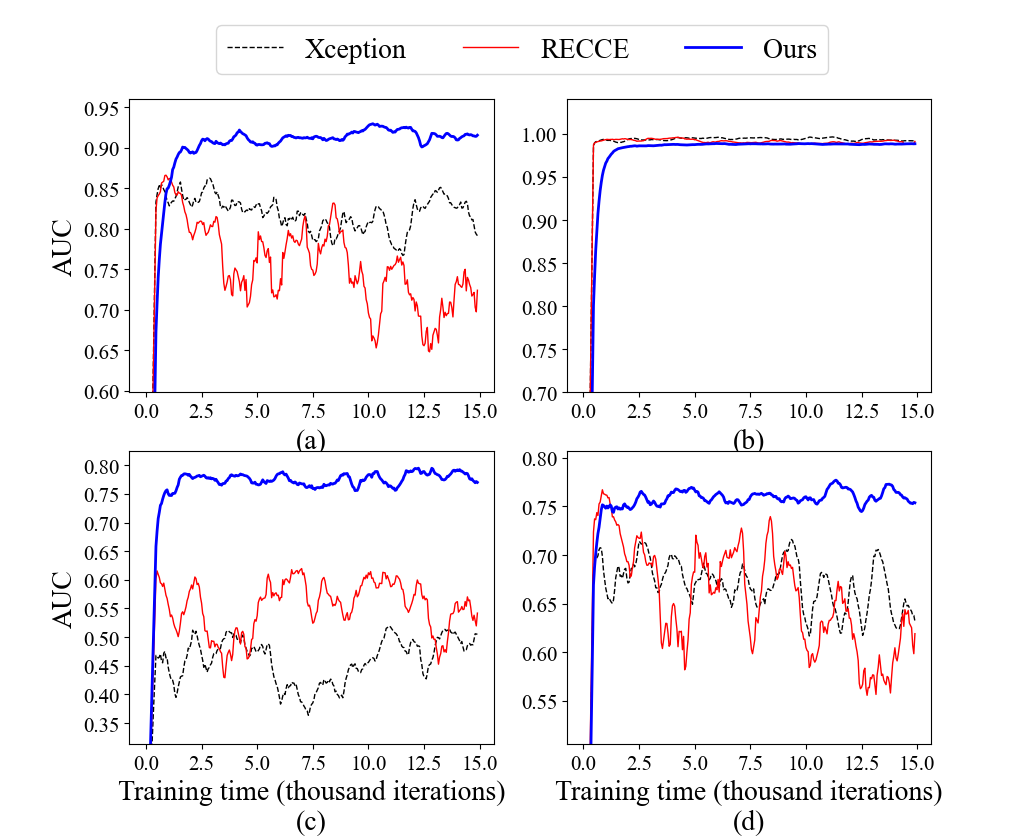}
  \vspace{-1.5em}
   \caption{Comparing the validation curves of RFFR-based deepfake detector and previous methods. Detectors are trained on the F2F subset of FF for $15k$ iterations and validated on four different subsets. (a) to (d) correspond to experiments on DF, F2F, FSW and NT.  Results are reported in AUC (\%). All three methods perform well when validated on F2F. However, under cross-manipulation settings, only our method avoids overfitting during training. The curves are smoothed for better visibility.}
\label{fig:validation-free}
\vspace{-1em}
\end{figure}

Models expected to generalize to other domains benefit from target domain validations~\cite{domainbed}. By frequently performing model validation, we can select the model  that best suits the detection of target manipulation, resulting in high performance on the test set. While using such an \textit{oracle} could be acceptable for the early development of cross-domain algorithms~\cite{domainbed}, it is not ideal for applications, as labeled data of unseen manipulation is usually not available. 

In this section, we demonstrate the potential of our deepfake detector to circumvent this practice and therefore avoid the need for extra validation data. As shown in \cref{tab:validation}, we train our classifier on F2F for 15k iterations and directly use the final model for testing. Simultaneously, we employ four validation sets to select the models with the best validation performances on target data. All validated and non-validated models are tested under the same conditions. We report all results on the target test sets in Table~\ref{tab:validation}. The performance gaps between validated and non-validated models are reported along with the test results. Although our non-validated models are not performing as well as those selected with a validation set, we show that our model remains effective on target data, with a maximum performance drop of $3.15\%$ and an average drop of $1.97\%$. However, previous methods~\cite{xception, recce} suffer from significantly larger performance drops when evaluated under the same procedure. 

To take a closer look at how the cross-manipulation performances vary during training, we train the deepfake detectors again with F2F. We test the AUC performances on all target subsets every 50 iterations to produce validation curves in \cref{fig:validation-free}. Our RFFR-based deepfake detector consistently maintains a high performance long after its peaks without serious overfitting. On the contrary, both previous methods compared here overfit quickly after reaching their highest target domain performances. In addition, compared methods exhibit large fluctuations across different evaluations, while our model remains stable. This suggests that with RFFR, our model focuses exclusively on generalizable features which fall outside the distribution of RFFR. Such resistance to overfitting guarantees our model a satisfying performance even when labeled validation sets are not available, which is generally expected in practice. We present more results on validation-free evaluations in Appendix.

\section{Conclusion}
\label{sec:conclusion}

In this paper, we propose Real Face Foundation Representation Learning (RFFR), which learns a general representation of real faces by training on a large number of real face images. The model is sensitive to unknown features outside the distribution of RFFR, and therefore effectively identifies fake face images of various forms. Extensive experiments under cross-domain settings demonstrate the superiority of performing deepfake detection with RFFR, as we show the resulting detectors are less prone to overfitting and generalize well to unseen face manipulations.

{\small
\bibliographystyle{ieee_fullname}
\bibliography{11_references}
}

\ifarxiv \clearpage \appendix
\label{sec:appendix}

\section*{\centering{Appendix}}
\section{Validation-free Evaluations}
\label{sec:more-validation-free}

We present the full results of validation-free evaluation for cross-manipulation deepfake detection. 
We ran the public code of Xception~\cite{xception}, RECCE~\cite{recce} and UIA-ViT~\cite{uia} to obtain their results under our validation-free setting. In specific, we train all models for exactly $15k$ iterations, which is long enough for them to reach peak validation performances and continue training for a while. We then use the final model to directly test on all subsets of FF. Results are reported in~\cref{tab:validation-free-cross-manipulation}.

We emphasize that this setting more closely resembles realistic scenarios, where validation sets are generally not available. Under this setting, we outperform the compared state-of-the-art methods by large margins of $7.46\%$, $8.64\%$ and $5.46\%$ when trained on DF, F2F and FSW respectively, and exhibit competitive performance on NT as well. Note that although we do not use validation sets, our results are not significantly weakened. We observe an only $2.05\%$ average decrease compared to evaluating with validation sets, which is reported in~\cref{tab:cross-manipulation} of the main paper. This comprehensive result demonstrates an impressive ability of RFFR-based deepfake detectors to avoid overfitting and be applied effectively for practical deepfake detection. 
\begin{table}[h]
\setlength\tabcolsep{4.5pt} 
\caption{Validation-free cross-manipulation performances in terms of AUC(\%). Classifiers are trained on one subset of FF and tested on all four subsets. \emph{No validation set is used for model selection.} Intra-domain results are marked in gray.}
\vspace{-1em}

\label{tab:validation-free-cross-manipulation}
\begin{center}  
\scalebox{0.8}{
\begin{tabular}{c|l|cccc|c}
\toprule
Training &\multirow{2}*{Method} & \multicolumn{4}{c|}{Test data} & \multirow{2}*{Avg} \\
\cmidrule(lr){3-6}
     data  &            ~                   & DF    & F2F   & FSW   & NT    & ~   \\
    \midrule
\multirow{4}*{DF}  & Xception~\cite{xception} &\cellcolor{Gray}\textbf{99.61} & 57.17 & 25.99 & 62.52 & 61.32 \\
       & RECCE~\cite{recce}       & \cellcolor{Gray}99.51 & 66.29 & 40.38 & \textbf{74.57} & 70.19 \\
       & UIA-ViT~\cite{uia}       & \cellcolor{Gray}99.37 & 62.86 & 54.54 & 65.10 & 70.47 \\
       & Ours                                  & \cellcolor{Gray}99.30  & \textbf{73.45} & \textbf{67.52} & 71.45 & \textbf{77.93} \\
\midrule
\multirow{4}*{F2F} & Xception~\cite{xception} & 83.08 & \cellcolor{Gray}99.12 & 46.63 & 64.93 & 73.44 \\
       & RECCE~\cite{recce}       & 74.51 & \cellcolor{Gray}99.22 & 50.17 & 59.46 & 70.84 \\
       & UIA-ViT~\cite{uia}        & 83.95 & \cellcolor{Gray}99.01 & 61.86 & 63.97 & 77.20 \\
       & Ours                                  & \textbf{91.56} & \cellcolor{Gray}\textbf{99.38} & \textbf{76.01} & \textbf{76.41} & \textbf{85.84} \\
\midrule
\multirow{4}*{FSW} & Xception~\cite{xception} & 53.31 & 57.48 & \cellcolor{Gray}\textbf{99.72} & 44.56 & 63.77 \\
       & RECCE~\cite{recce}       & 49.85 & 65.77 & \cellcolor{Gray}99.68 & 55.95 & 67.81 \\
       & UIA-ViT~\cite{uia}        & 79.33 & 65.60 & \cellcolor{Gray}99.23 & 50.90 & 73.77 \\
       & Ours                                  & \textbf{85.24} & \textbf{75.14} & \cellcolor{Gray}99.49 & \textbf{57.06} & \textbf{79.23} \\
\midrule
\multirow{4}*{NT}  & Xception~\cite{xception} & \textbf{90.83} & 68.68 & 38.45 &\cellcolor{Gray}\textbf{97.11} & 73.77 \\
       & RECCE~\cite{recce}       & 86.98 & 72.20 & \textbf{51.10} & \cellcolor{Gray}97.06 & \textbf{76.84} \\
       & UIA-ViT~\cite{uia}        & 78.98 & 64.80 & 44.55 & \cellcolor{Gray}95.62 & 70.99 \\
       & Ours                                  & 81.09 & \textbf{73.59} & 50.40  & \cellcolor{Gray}95.84 & 75.23 \\
\bottomrule
\end{tabular}}
\vspace{-1em}
\end{center}
\end{table}

\section{Additional Ablation Study}
\subsection{Ablation Study for Residuals}
\label{sec:ablation_residuals}

We use a pretrained MAE to obtain residual images that signal potential artifacts. In this section, we perform ablation study for the residuals, where we compare our method with other residual generation techniques. All models are trained on F2F and tested on FSW. As a baseline, we train a single-branch ViT that learns without any residuals, and only accepts random original image blocks. Subsequently, we train three dual-branch ViTs that accepts different residuals. In addition to our residual generated by MAE, we propose two other options of residual generation. We present the results in \cref{tab:unet}.

As mentioned in the paper, we use a UNet-based autoencoder trained with real faces to produce similar residual images by subtracting the reconstructed images from the originals. We show that these residual blocks cause a slight decrease ($-0.39\%$) in generalization performance. This is likely due to perfect reconstructions that leave both real and fake residual images with no information to exploit.

We also explore high-pass filters, another potential source of residual images. In some cases, directly applying high-pass filters on deepfake images yields visually similar results to our MIM-based residuals. However, they tend to treat all image regions equally and fail to expose artifacts that are distinct from the rest of the image. In our experiment, we show that while high-pass filtered images improve upon the baseline, the improvement is marginal ($0.79\%$) compared to our MIM-based residuals.

Finally, we show that our design of MIM-based residuals brings about a $7.86\%$ improvement in performance, significantly outperforming compared residual generation methods. As demonstrated in the main paper, the MIM-based method effectively differentiates between the processing of real and fake samples and successfully highlights potential forgery patterns in its residuals. Therefore, it makes a substantial contribution to the generalization performance of deepfake detectors.

\begin{table}[]
\setlength\tabcolsep{4.5pt} 
\caption{Deepfake detection performances of different residuals. Classifiers are trained on F2F and tested on FSW. We present the detection results in AUC (\%). }
\vspace{-1em}
\label{tab:unet}
\begin{center}
\begin{tabular}{c | c | c}
\toprule
Training data & Residuals & Test AUC (\%)\\
\midrule
\multirow{4}*{F2F} & None & 70.76 \\
& Autoencoder & 70.37 \\
& High-pass filter & 71.55 \\
& MIM (Ours) & \textbf{78.62} \\
\bottomrule
\end{tabular}
\vspace{-1em}

\end{center}
\end{table}

\subsection{Ablation Study for Block Sizes}
\label{sec:ablation_block}

To perform masked image modeling, we split each image into $k\times k$ blocks and inpaint one block at a time. Selecting an appropriate block size requires balancing performance and efficiency. Large blocks hinder deepfake detection with increased noise due to the difficulty to accurately inpaint. Small blocks cause longer inference time with more forward passes required to complete a reconstruction. To strike a balance, we opt for $k = 4$ for optimal detection performance and efficient inference. We present deepfake detection results of different block sizes in~\cref{tab:block}. Note that smaller blocks does not improve detection, but significantly prolongs inference time.

\begin{table}[!h]
\setlength\tabcolsep{4.5pt} 
\caption{Comparing performances and inference time of different block sizes. Models trained on F2F. Results in AUC(\%)}
\label{tab:block}
\vspace{-1em}
\begin{center}
\scalebox{0.8}{
\begin{tabular}{c|ccccc|c}
\toprule
Split &  DF & F2F & FSW & NT & Avg & Inference Time \\ 
\midrule
$2 \times 2$ & $91.44$ & $98.07$ & $74.14$ & $75.67$ & $84.83$ & $4\times$ MAE Inference \\ 
$4 \times 4$ & $\textbf{93.44}$ & $\textbf{99.61}$ & $78.62$ & $\textbf{79.56}$ & $\textbf{87.81}$ & $16 \times$ MAE Inference\\
$6 \times 6$ & $93.08$ & $99.18$ & $\textbf{79.34}$ & $78.58$ & $87.55$ & $36 \times$ MAE Inference\\
\bottomrule
\end{tabular}}
\vspace{-1em}
\end{center}
\end{table}

 \fi

\end{document}